\documentclass[%
 aip,
 jmp,%
 amsmath,amssymb,
 reprint,%
]{revtex4-1}

\usepackage{graphicx}
\graphicspath{{figures_pdf/}}
\usepackage{dcolumn}
\usepackage{bm}
\usepackage{subfigure}

\usepackage{soul}
\usepackage{amsmath,amssymb,amsfonts}
\usepackage{graphics}
\usepackage{color}
\usepackage{xcolor}
\definecolor{rev1}{rgb}{0,0,0}

\usepackage{algorithm}
\usepackage{algorithmic}
\usepackage{enumitem}
\setlist[itemize]{leftmargin=*}

\usepackage{comment}
\usepackage{hyperref}
\usepackage{lipsum}
\usepackage{tabularx}
\usepackage{mathrsfs}
\usepackage{stackengine}

\begin{document}

\title{Physics guided machine learning using simplified theories}

\author{Suraj Pawar}
\affiliation{ 
School of Mechanical \& Aerospace Engineering, Oklahoma State University, Stillwater, OK 74078, USA.
}%

\author{Omer San}%
 \email{osan@okstate.edu}
\affiliation{ 
School of Mechanical \& Aerospace Engineering, Oklahoma State University, Stillwater, OK 74078, USA.
}%

\author{Burak Aksoylu}%
\affiliation{ 
	Texas A\&M University-San Antonio, Department of Mathematical, Physical, and Engineering Sciences, San Antonio, TX 78224, USA.
}%

\author{Adil Rasheed}%
\affiliation{ 
	Department of Engineering Cybernetics, Norwegian University of Science and Technology, 7465 Trondheim, Norway.
}%
\affiliation{
Department of Mathematics and Cybernetics, SINTEF Digital, 7034 Trondheim, Norway.
}%
\author{Trond Kvamsdal}%
\affiliation{
Department of Mathematical Sciences, Norwegian University of Science and Technology, 7491 Trondheim, Norway.
}%
\affiliation{
Department of Mathematics and Cybernetics, SINTEF Digital, 7034 Trondheim, Norway.
}%


\date{\today}

\begin{abstract}
Recent applications of machine learning, in particular deep learning, motivate the need to address the generalizability of the statistical inference approaches in physical sciences. In this letter, we introduce a modular physics guided machine learning framework to improve the accuracy of such data-driven predictive engines. The chief idea in our approach is to augment the knowledge of the simplified theories with the underlying learning process. To emphasise on their physical importance, our architecture consists of adding certain features at intermediate layers rather than in the input layer. To demonstrate our approach, we select a canonical airfoil aerodynamic problem with the enhancement of the potential flow theory. We include features obtained by a panel method that can be computed efficiently for an unseen configuration in our training procedure. By addressing the generalizability concerns, our results suggest that the proposed feature enhancement approach can be effectively used in many scientific machine learning applications, especially for the systems where we can use a theoretical, empirical, or simplified model to guide the learning module.       



\end{abstract}


\maketitle



Advanced machine learning (ML) models like deep neural networks are powerful tools for finding patterns in complicated datasets. However, as these \emph{universal approximators}\cite{hornik1989multilayerua,hagan1994training,hagan1997neural,vidal2017mathematics} become complex, the number of trainable parameters (weights) quickly explodes, adversely affecting their interpretability, and hence their trustworthiness.  Using these models in combination with other traditional models compromises the trustworthiness of the overall system. 
\textcolor{rev1}{While such techniques have been emerging in both scientific computing and ML fields \cite{krasnopolsky2006complex,lee1997hybrid,tang2001coupling,marsland2002self,raissi2019physics,zhu2019physics,swischuk2019projection,sekar2019fast,beck2019deep,rahman2018hybrid,maulik2019sub,maulik2019subgrid,pan2020physics,mohan2020embedding,muralidhar2020physics,qian2020lift,pawar2020data,ahmed2020long,ahmed2020nudged,pawar2020long,pawar2020interface,pawar2020prf,ahmed2020interface}, they offer many opportunities to fuse topics in domain-specific knowledge, numerical linear algebra, and theoretical computer science \cite{golub2006bridging} towards improving the generalizability of these models.}    
In this letter, we focus on a modular approach for improving generalizability by augmenting input features with simplified physics-based models or theories. By the term of \emph{generalizability}, we refer to the performance of the learning engine for unseen configurations. Such new situations are pervasive in scientific applications. In many cases, there is a grand challenge in making accurate predictions using data-driven methods based strictly on statistical inference. For example, in fluid mechanics, potential flow and boundary layer theories often provide simplistic models pertaining to a certain level of approximations. A remaining task is how to combine these models for making more generalizable learning architectures. Therefore, our approach provides intermediate features based on approximate physics to statistical models that can relate or fulfill the gap between these simplified theories and more accurate models (i.e., experimental or high fidelity computational fluid dynamics simulation data sets).        


\begin{figure*}[htbp]
\centering
\includegraphics[width=0.99\textwidth]{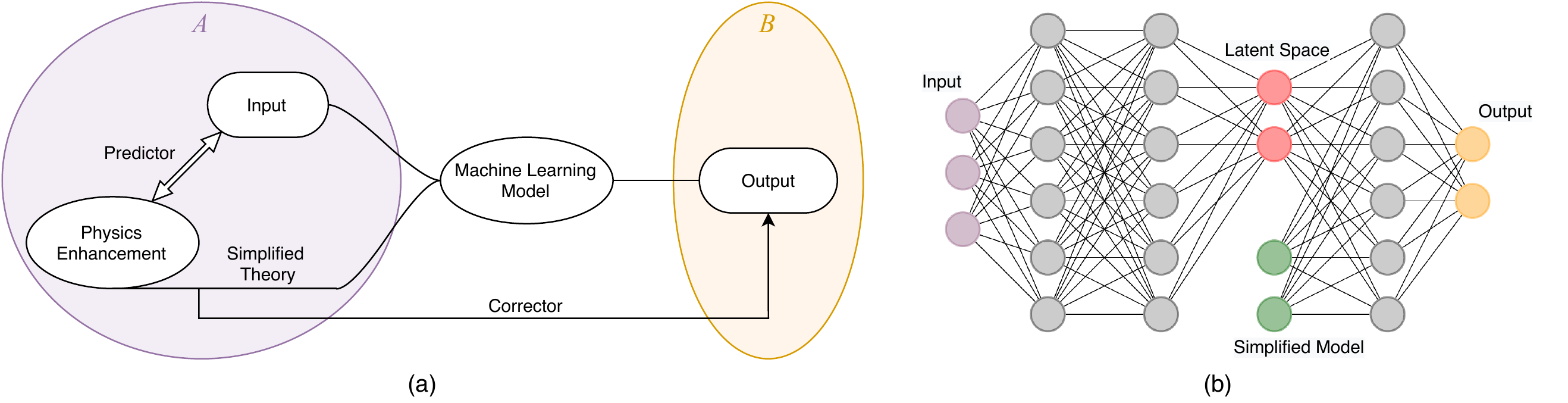}
\caption{Physics guided machine learning (PGML) framework to train a learning engine between processes $A$ and $B$: (a) a conceptual PGML framework \textcolor{rev1}{which shows different ways of incorporating physics into machine learning models. The physics can be incorporated using feature enhancement of the ML model based on the domain knowledge, embedding simplified theories directly into ML models, and corrector approach in which the output of the ML model is constrained using the governing equations of the system}, 
and (b) an overview of the typical neural network architecture for the PGML framework.}
\label{fig:pgml}
\end{figure*}

A wide variety of engineering tasks such as optimal control, design optimization, uncertainty quantification requires the prediction of the quantity of interest in real-time. In such scenarios, the partial differential equation (PDE) based discretization methods are not feasible as they can take days or weeks to simulate different configurations. Reduced order models (ROMs) are the state-of-the-art models that construct the basis from the past data, and then solve the governing equations after projecting them on the low dimensional manifold \cite{lucia2004reduced,taira2019modal,fonn2019fast}.
The main limitation of such projection based intrusive ROMs is that it requires a complete description of the dynamics of the system, and often this information is unavailable or inadequate for the desired application. Recently model-free prediction using machine learning has proven successful for many physical systems \cite{pathak2018model,white2019neural,geneva2020modeling}. 
One of the main challenges with these non-intrusive models is the prediction for unseen data and its interpretation. Even though probabilistic machine learning methods like Bayesian neural networks can give the uncertainty bound on the predicted state \cite{zhu2019physics,maulik2020probabilistic}, the generalizability of non-intrusive models is not on par with physics-based models. On the other hand, the simplified models like the lumped-capacitance model for heat transfer analysis, and Blasius boundary layer model in fluid mechanics are highly generalizable to different conditions. Therefore, it is important to leverage the knowledge of physical systems into learning, and to this end, we make progress towards physics guided machine learning by embedding features from simplified physical models into the neural network architecture. The proposed framework is flexible enough to be applied to many physical systems and it has great potential in scientific machine learning.

We now introduce different components of the physics guided machine learning (PGML) framework as depicted in Figure~\ref{fig:pgml}(a). In supervised machine learning, the input vector $\mathbf{x} \in \mathbb{R}^m$ is fed to the machine learning model (for example, the neural network in our case), and the \textcolor{rev1}{mapping from the input vector to} output vector $\mathbf{y} \in \mathbb{R}^n$ is learned \textcolor{rev1}{through training}. The neural network is trained to learn the function $F_\theta$, parameterized by $\theta$, that includes the weights and biases of each neuron. The function $F_\theta$ should be such that it approximates the known labels and the cost function is minimized. Usually, for the regression problems, the cost function is the mean squared error between true and predicted output, i.e., $C(\mathbf{x},\theta)=||\mathbf{y} -F_{\theta}(\mathbf{x})||_2$. The weights and biases of the neural network are optimized using any gradient-descent algorithm like stochastic gradient descent. In the PGML framework, the neural network is augmented with the output of the simplified physics-based model. The features extracted from simplified physics-based models can be either combined with input features, or they can be merged into hidden layers along with latent variables. During the training, the PGML framework is provided with $(\mathbf{x}, G(\mathbf{x}))$, where $G$ is the model based on simplified theories for the problem at hand, and the parameters $\theta$ are optimized based on the true output $\mathbf{y}$. The features from the simplified model $G$ can also be fed at an intermediate hidden layer along with learned latent representation. 

The PGML framework allows us to extract the knowledge from the simplified physics-based model and combine it with the latent variables of the system discovered by the neural network at intermediate hidden layers. This information from the physics-based model assists the neural network in constraining its output to a manifold of physically realizable models. It encourages the neural network to learn the physically consistent representation of the quantity of interest drawn from complex distributions such as pressure and velocity field of fluid simulations. Another advantage of the PGML framework is that it brings interpretability to otherwise black-box models. We highlight here that the PGML framework allows us to incorporate the physics of the problem even during the prediction stage, and not just the training as in the case of approaches like regularization based on governing equations. 

In a nutshell, we highlight that there might be a handful of simplified models (e.g., similarity solutions, lumped parameter models, zone models, zero or one-dimensional models, etc) to approximate or describe the underlying physical processes in many disciples. For example, in aerodynamics, the use of simplistic panel methods is a proven approach for analysis of inviscid flow over bodies, especially for a smaller angle of attacks where the potential flow theory becomes valid. The execution time for these simplified models is significantly smaller compared to the run time needed for full-fledged CFD simulations. Therefore, in the PGML method, we propose to fuse the knowledge coming from such simplified theories in a statistical inference architecture. This is accomplished by a feature enhancement procedure as described in Figure~\ref{fig:pgml}(b). Specifically, it constitutes a predictor-corrector philosophy where we first run a simplified theory (computationally less demanding) to estimate an intermediate prediction as an input for the overall ML architecture. We then estimate the quantity of interest based on the enhancement procedure abstracted by $G(\mathbf{x})$. We hypothesize that the predictive performance and generalizability of statistical inference will be significantly improved by the involvement of $G(\mathbf{x})$ as illustrated with the aerodynamic performance prediction task.                   

\emph{Results and Discussion} --- We demonstrate the PGML framework using the aerodynamic performance prediction task. This is a problem relevant to many applications such as predictive control\cite{zha2007high}, and design optimization\cite{legresley2000airfoil}. The prediction of flow around an airfoil is a high-dimensional, multiscale, and nonlinear problem that can be solved using high-fidelity methods like computational fluid dynamics (CFD). However, these methods are computationally intractable as the design space exploration increases. In certain flow regimes, the simplified methods like panel codes come with a non-negligible difference between the actual dynamics and approximate models for real-world problems. The full-order CFD simulations typically take extensive computational run time, thus limiting their use in many inverse modelling methodologies that require a model run to be performed in each iteration. To overcome these challenges, combining CFD models with machine learning to build a data-driven surrogate model is gaining widespread popularity\cite{zhang2018application,bhatnagar2019prediction,rajaram2020deep}. In this example, we exploit the relevant physics-based features from panel methods through the PGML framework to enhance the generalizability of data-driven surrogate models.  

The training data for the neural network is generated using a series of numerical simulations. The main goal of this work is to demonstrate how a simplified model can be used in the PGML framework and therefore, to reduce the computational complexity, we utilize XFOIL\cite{drela1989xfoil} for the aerodynamic analysis of different airfoils instead of full CFD simulation to get the most accurate results for the forces on the airfoil. The lift coefficient data were obtained for different Reynolds numbers between $1 \times 10^6$ to $4 \times 10^6$ and several angles of attacks in the range of $-20$ to $+20$. A total of 168 sets of two-dimensional airfoil geometry were generated for training the neural network. This includes NACA 4-digit, NACA210, NACA220, and NACA250 series, and each airfoil is represented by 201 points. The maximum thickness of all airfoils in the training dataset was between 6\% to 18\% of the chord length. We use the NACA23012 and NACA23024 airfoil geometry as the test dataset to evaluate the predictive capability of the trained neural network. The simplified model used to generate the physics-based feature corresponds to the Hess-Smith panel method\cite{hess1990panel} based on potential flow theory. We highlight that our testing airfoils are selected not only from a different NACA230 series (i.e., not used in the training dataset) but also the maximum thickness of 24\% is well beyond the thickness ratio limit included in our training data set.   

\begin{figure}[htbp]
\centering
\includegraphics[width=0.49\textwidth]{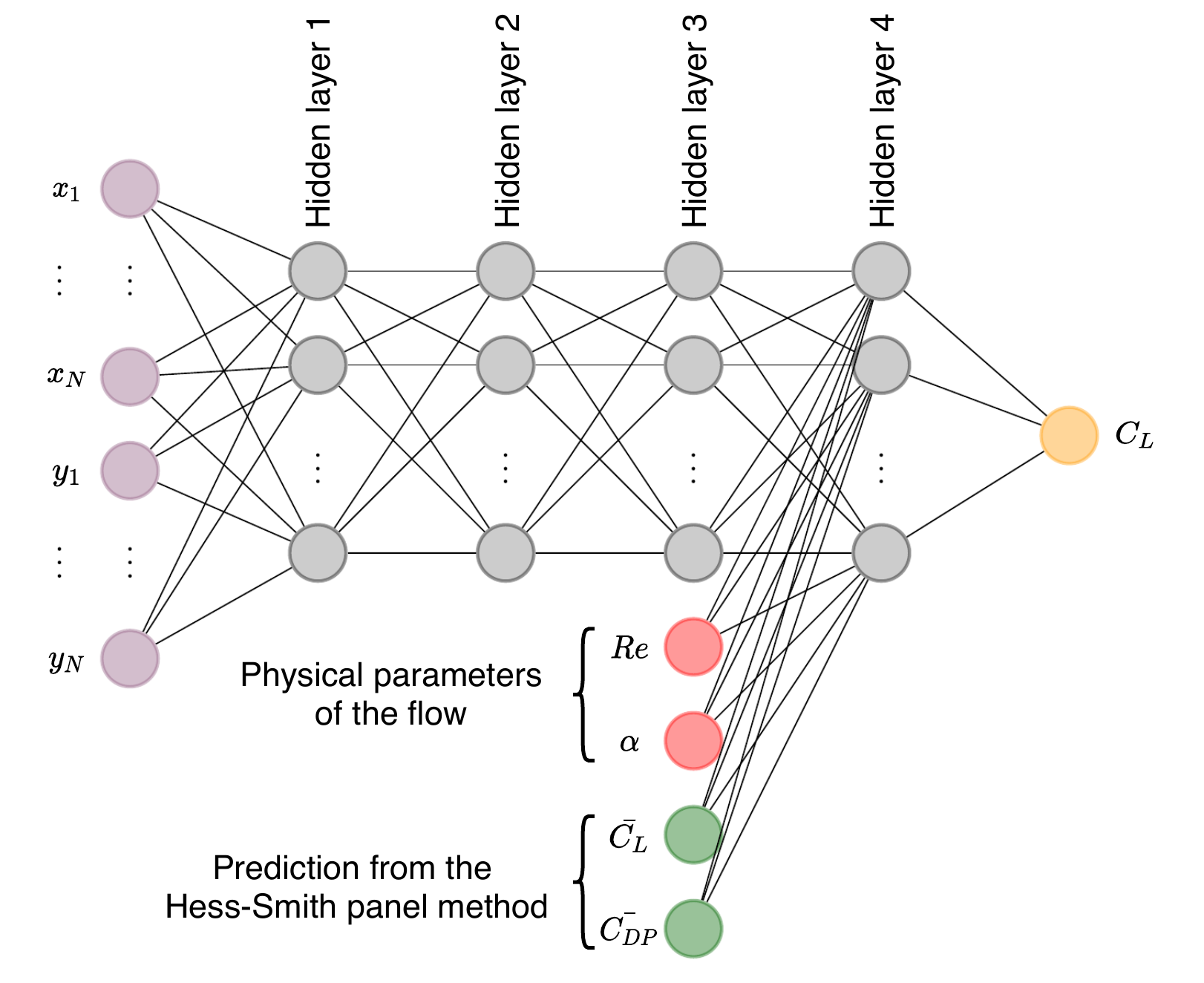}
\caption{\textcolor{rev1}{The representative neural network architecture of the PGML framework used in this study for aerodynamic forces prediction task. The latent variables at the third hidden layers are augmented with the physical parameters of the flow (i.e., the Reynolds number and angle of attack) and the prediction from the Hess-Smith panel method (i.e., lift coefficient and pressure drag coefficient denoted as $\bar{C_L}$ and $\bar{C_{DP}}$, respectively).} }
\label{fig:pgsi_airfoil_nn}
\end{figure}

\begin{figure}[htbp]
\centering
\includegraphics[width=0.49\textwidth]{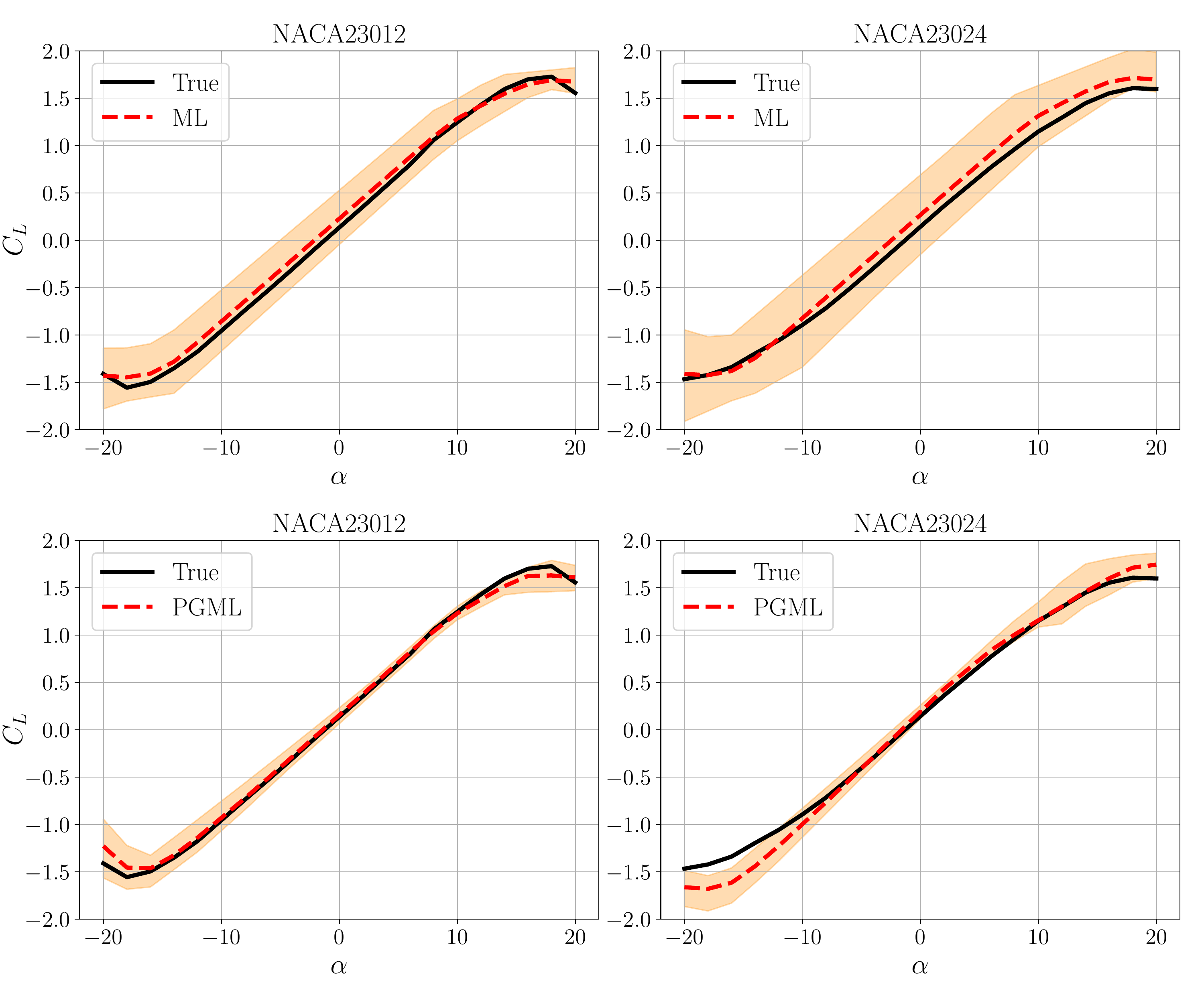}
\caption{Actual versus predicted lift coefficient ($C_L$) for NACA23012 and NACA23024 airfoils at $Re=3 \times 10^6$ using ML and PGML framework. The dashed red curve represent the average of the predicted lift coefficient by all ML models (i.e., testing runs with different initialization seeds). Both airfoil geometries were not included in the training. Here we note that there is a significant reduction of uncertainty in performance when we use the proposed PGML framework, especially for the smaller angle of attacks where the embedded simplistic physics-based model is valid. The physics embedding in these tests are based on the utilization of the Hess-Smith panel method, which is limited mostly for the angle of attacks between -10 and +10 degrees, which further verifies the success of the proposed PGML framework.}
\label{fig:airfoil_nn}
\end{figure}

The neural network architecture used in this study has four hidden layers with 20 neurons in each hidden layer. The physical parameters, i.e., the Reynolds number and the angle of attack are concatenated at the third hidden layer along with the latent variables at that layer. In the PGML model, we augment the latent variables at the third layer with the lift coefficient and the pressure drag coefficient predicted by the panel method along with physical parameters of the flow (i.e., the Reynolds number and angle of attack). Therefore, the third layer of the PGML neural network has 24 latent variables. \textcolor{rev1}{The representative neural network for the PGML framework to predict the aerodynamic forces on an airfoil is displayed in Figure~\ref{fig:pgsi_airfoil_nn}.} 
We utilize ensemble of neural networks to quantify the predictive uncertainty\cite{tibshirani1996comparison,heskes1997practical,lakshminarayanan2017simple}. In this method, a small number of neural networks are trained from different initialization and the variance of the ensemble's prediction is interpreted as the epistemic uncertainty. This method is appealing due to its simplicity, scalability, and strong empirical results of the uncertainty estimate that are as good as the Bayesian neural networks\cite{lakshminarayanan2017simple}. We train multiple neural network models using different values of the initial weights and biases. The weights and biases of each model are initialized using the Glorot uniform initializer and different random seed numbers are used to ensure that the different values of weights and biases are assigned for each model. The ensemble of all these models indicates the model uncertainty estimate of the predicted lift coefficient. Figure~\ref{fig:airfoil_nn} shows the actual and predicted lift coefficient for the NACA23012 and NACA23024 airfoil geometry. The reference \emph{True} performance is obtained by XFOIL. The ML corresponds to a simple feed-forward neural network that uses the airfoil $x$ and $y$ coordinates as the input features, \textcolor{rev1}{and the physical parameters of the flow are concatenated at the third hidden layer along with the latent variables at that layer.} 

As shown in Figure~\ref{fig:airfoil_nn}, we can see that the uncertainty in the prediction of the lift coefficient by the PGML model is less than the ML model for both NACA23012 and NACA23024 airfoils. The proposed PGML framework provides significantly more accurate predictions with reduced uncertainty for the angle of attacks between -10 and +10 degrees. This further illustrates the viability of the proposed PGML framework, since the physics embedding considered here employs constant source panels and a single vortex to approximate the potential flow around the airfoil. We can also notice that the uncertainty is higher for the angle of attacks outside the range of -10 to +10 degrees. This finding is not surprising as the Hess-Smith panel method is a proven method for analysis of inviscid flow over airfoil for the smaller angle of attacks regime. We highlight that the maximum thickness of an airfoil included in the training dataset is 18\% of the chord length. Therefore, the uncertainty in the prediction of the lift coefficient by the ML model is higher for the NACA23024 airfoil compared to the NACA23012 airfoil. The inclusion of physics-based features from the panel method in the PGML model leads to a reduction in this uncertainty estimate. These results clearly show the advantage of augmenting the neural network model with simplified theories and the potential of the PGML framework for physical systems. \textcolor{rev1}{One of the important caveats with any neural network is its design and the hyperparameters. The neural architecture search and hyperparameter optimization are important processes for the success of the PGML framework. If the network is shallow, then it has less expressive capabilities and that deteriorates the prediction. On the other hand, if the neural network architecture is very deep and there is no sufficient training data, then the network fits the training data very well. However, its generalizability is reduced and this is usually indicated by an increase in the loss of the validation dataset after a few iterations of the training. Some of the strategies to mitigate overfitting issues in the deep neural network are using $L_1/L_2$ regularization, applying dropout, early stopping, data augmentation, and $k-$fold cross-validation. Overall, our findings suggest that the absorption of physical information into ML methods improves the modeling uncertainty of the ML architectures.}          


\emph{Concluding Remarks} --- \textcolor{rev1}{The data-driven models derived from the data collected from satellite measurements, internet of things (IoT) devices, and numerical simulation are increasingly being applied for scientific applications in fluid dynamics. While these data-driven models are successful, it might be vital to complement them with physical laws that have been studied for many decades. To this end, physics-informed machine learning approaches such as embedding soft and hard constraints designed based on governing laws of the physical system,  have been proposed \cite{raissi2019physics,zhu2019physics,pan2020physics,mohan2020embedding}.} In this work, we illustrated how physics-based models derived from simplified approximations of the physical system can be coupled within data-driven models like neural networks. The PGML framework introduced in this study allows us to take advantage of the generalizability of physics-based simplistic models and the robustness of data-driven models. We demonstrated the proof-of-concept for the aerodynamic performance analysis of airfoil geometry. The physics-based features computed from the simplistic panel method are coupled with the latent representation learned at the intermediate layers of the neural network. The inclusion of these physics-based features assists the neural network model in reducing the uncertainty of the lift coefficient prediction. Therefore, the PGML framework is successful in improving the generalizability of data-driven models. 

Additionally, the PGML framework will also be useful for physical systems where the data is scarce. For example, the generation of training data is computationally expensive when dealing with larger problems such as the optimization of a wind farm layout \cite{samorani2013wind} or the shape optimization of a three-dimensional wing \cite{epstein2009comparative}. The simplified models like the actuator disk theory can be used along with a velocity field generated from few high-fidelity numerical simulations to build a physics-guided neural network-based surrogate model of the wind farm that can be coupled with any optimization algorithm. Another area where the PGML framework may bear huge potential is the digital twin technologies for physical systems\cite{rasheed2020digital}, where the digital replica of the physical system is build for tasks such as real-time control, efficient operation, and accurate forecast of maintenance schedules. \textcolor{rev1}{In the present study, we started with XFOIL to assess the feasibility of the proposed framework. Despite its simplicity, this is the first step towards demonstrating the PGML framework for problems in fluid dynamics.} In our future work we plan to extend the PGML approach to complex and high-dimensional problems like prediction of terrain induced atmospheric boundary layer flows or flows around bluff bodies immersed in fluid to illustrate the true capability of the PGML framework.

This material is based upon work supported by the U.S. Department of Energy, Office of Science, Office of Advanced Scientific Computing Research under Award Number DE-SC0019290. O.S. gratefully acknowledges their support. 

The data that supports the findings of this study are available within the article. Implementation details and Python scripts can be accessed from the Github repository\cite{githubpgml}.

\bibliography{references}

\begin{thebibliography}{49}%
\makeatletter
\providecommand \@ifxundefined [1]{%
 \@ifx{#1\undefined}
}%
\providecommand \@ifnum [1]{%
 \ifnum #1\expandafter \@firstoftwo
 \else \expandafter \@secondoftwo
 \fi
}%
\providecommand \@ifx [1]{%
 \ifx #1\expandafter \@firstoftwo
 \else \expandafter \@secondoftwo
 \fi
}%
\providecommand \natexlab [1]{#1}%
\providecommand \enquote  [1]{``#1''}%
\providecommand \bibnamefont  [1]{#1}%
\providecommand \bibfnamefont [1]{#1}%
\providecommand \citenamefont [1]{#1}%
\providecommand \href@noop [0]{\@secondoftwo}%
\providecommand \href [0]{\begingroup \@sanitize@url \@href}%
\providecommand \@href[1]{\@@startlink{#1}\@@href}%
\providecommand \@@href[1]{\endgroup#1\@@endlink}%
\providecommand \@sanitize@url [0]{\catcode `\\12\catcode `\$12\catcode
  `\&12\catcode `\#12\catcode `\^12\catcode `\_12\catcode `\%12\relax}%
\providecommand \@@startlink[1]{}%
\providecommand \@@endlink[0]{}%
\providecommand \url  [0]{\begingroup\@sanitize@url \@url }%
\providecommand \@url [1]{\endgroup\@href {#1}{\urlprefix }}%
\providecommand \urlprefix  [0]{URL }%
\providecommand \Eprint [0]{\href }%
\providecommand \doibase [0]{http://dx.doi.org/}%
\providecommand \selectlanguage [0]{\@gobble}%
\providecommand \bibinfo  [0]{\@secondoftwo}%
\providecommand \bibfield  [0]{\@secondoftwo}%
\providecommand \translation [1]{[#1]}%
\providecommand \BibitemOpen [0]{}%
\providecommand \bibitemStop [0]{}%
\providecommand \bibitemNoStop [0]{.\EOS\space}%
\providecommand \EOS [0]{\spacefactor3000\relax}%
\providecommand \BibitemShut  [1]{\csname bibitem#1\endcsname}%
\let\auto@bib@innerbib\@empty
\bibitem [{\citenamefont {Hornik}, \citenamefont {Stinchcombe},\ and\
  \citenamefont {White}(1989)}]{hornik1989multilayerua}%
  \BibitemOpen
  \bibfield  {author} {\bibinfo {author} {\bibfnamefont {K.}~\bibnamefont
  {Hornik}}, \bibinfo {author} {\bibfnamefont {M.}~\bibnamefont {Stinchcombe}},
  \ and\ \bibinfo {author} {\bibfnamefont {H.}~\bibnamefont {White}},\
  }\bibfield  {title} {\enquote {\bibinfo {title} {Multilayer feedforward
  networks are universal approximators},}\ }\href@noop {} {\bibfield  {journal}
  {\bibinfo  {journal} {Neural Networks}\ }\textbf {\bibinfo {volume} {2}},\
  \bibinfo {pages} {359--366} (\bibinfo {year} {1989})}\BibitemShut {NoStop}%
\bibitem [{\citenamefont {Hagan}\ and\ \citenamefont
  {Menhaj}(1994)}]{hagan1994training}%
  \BibitemOpen
  \bibfield  {author} {\bibinfo {author} {\bibfnamefont {M.~T.}\ \bibnamefont
  {Hagan}}\ and\ \bibinfo {author} {\bibfnamefont {M.~B.}\ \bibnamefont
  {Menhaj}},\ }\bibfield  {title} {\enquote {\bibinfo {title} {Training
  feedforward networks with the marquardt algorithm},}\ }\href@noop {}
  {\bibfield  {journal} {\bibinfo  {journal} {IEEE Transactions on Neural
  Networks}\ }\textbf {\bibinfo {volume} {5}},\ \bibinfo {pages} {989--993}
  (\bibinfo {year} {1994})}\BibitemShut {NoStop}%
\bibitem [{\citenamefont {Hagan}, \citenamefont {Demuth},\ and\ \citenamefont
  {Beale}(1997)}]{hagan1997neural}%
  \BibitemOpen
  \bibfield  {author} {\bibinfo {author} {\bibfnamefont {M.~T.}\ \bibnamefont
  {Hagan}}, \bibinfo {author} {\bibfnamefont {H.~B.}\ \bibnamefont {Demuth}}, \
  and\ \bibinfo {author} {\bibfnamefont {M.}~\bibnamefont {Beale}},\
  }\href@noop {} {\emph {\bibinfo {title} {Neural network design}}}\ (\bibinfo
  {publisher} {PWS Publishing Co., Boston},\ \bibinfo {year}
  {1997})\BibitemShut {NoStop}%
\bibitem [{\citenamefont {Vidal}\ \emph {et~al.}(2017)\citenamefont {Vidal},
  \citenamefont {Bruna}, \citenamefont {Giryes},\ and\ \citenamefont
  {Soatto}}]{vidal2017mathematics}%
  \BibitemOpen
  \bibfield  {author} {\bibinfo {author} {\bibfnamefont {R.}~\bibnamefont
  {Vidal}}, \bibinfo {author} {\bibfnamefont {J.}~\bibnamefont {Bruna}},
  \bibinfo {author} {\bibfnamefont {R.}~\bibnamefont {Giryes}}, \ and\ \bibinfo
  {author} {\bibfnamefont {S.}~\bibnamefont {Soatto}},\ }\bibfield  {title}
  {\enquote {\bibinfo {title} {Mathematics of deep learning},}\ }\href@noop {}
  {\bibfield  {journal} {\bibinfo  {journal} {arXiv preprint arXiv:1712.04741}\
  } (\bibinfo {year} {2017})}\BibitemShut {NoStop}%
\bibitem [{\citenamefont {Krasnopolsky}\ and\ \citenamefont
  {Fox-Rabinovitz}(2006)}]{krasnopolsky2006complex}%
  \BibitemOpen
  \bibfield  {author} {\bibinfo {author} {\bibfnamefont {V.~M.}\ \bibnamefont
  {Krasnopolsky}}\ and\ \bibinfo {author} {\bibfnamefont {M.~S.}\ \bibnamefont
  {Fox-Rabinovitz}},\ }\bibfield  {title} {\enquote {\bibinfo {title} {Complex
  hybrid models combining deterministic and machine learning components for
  numerical climate modeling and weather prediction},}\ }\href@noop {}
  {\bibfield  {journal} {\bibinfo  {journal} {Neural Networks}\ }\textbf
  {\bibinfo {volume} {19}},\ \bibinfo {pages} {122--134} (\bibinfo {year}
  {2006})}\BibitemShut {NoStop}%
\bibitem [{\citenamefont {Lee}\ and\ \citenamefont {Oh}(1997)}]{lee1997hybrid}%
  \BibitemOpen
  \bibfield  {author} {\bibinfo {author} {\bibfnamefont {J.-W.}\ \bibnamefont
  {Lee}}\ and\ \bibinfo {author} {\bibfnamefont {J.-H.}\ \bibnamefont {Oh}},\
  }\bibfield  {title} {\enquote {\bibinfo {title} {{Hybrid learning of mapping
  and its Jacobian in multilayer neural networks}},}\ }\href@noop {} {\bibfield
   {journal} {\bibinfo  {journal} {Neural Computation}\ }\textbf {\bibinfo
  {volume} {9}},\ \bibinfo {pages} {937--958} (\bibinfo {year}
  {1997})}\BibitemShut {NoStop}%
\bibitem [{\citenamefont {Tang}\ and\ \citenamefont
  {Hsieh}(2001)}]{tang2001coupling}%
  \BibitemOpen
  \bibfield  {author} {\bibinfo {author} {\bibfnamefont {Y.}~\bibnamefont
  {Tang}}\ and\ \bibinfo {author} {\bibfnamefont {W.~W.}\ \bibnamefont
  {Hsieh}},\ }\bibfield  {title} {\enquote {\bibinfo {title} {Coupling neural
  networks to incomplete dynamical systems via variational data
  assimilation},}\ }\href@noop {} {\bibfield  {journal} {\bibinfo  {journal}
  {Monthly Weather Review}\ }\textbf {\bibinfo {volume} {129}},\ \bibinfo
  {pages} {818--834} (\bibinfo {year} {2001})}\BibitemShut {NoStop}%
\bibitem [{\citenamefont {Marsland}, \citenamefont {Shapiro},\ and\
  \citenamefont {Nehmzow}(2002)}]{marsland2002self}%
  \BibitemOpen
  \bibfield  {author} {\bibinfo {author} {\bibfnamefont {S.}~\bibnamefont
  {Marsland}}, \bibinfo {author} {\bibfnamefont {J.}~\bibnamefont {Shapiro}}, \
  and\ \bibinfo {author} {\bibfnamefont {U.}~\bibnamefont {Nehmzow}},\
  }\bibfield  {title} {\enquote {\bibinfo {title} {A self-organising network
  that grows when required},}\ }\href@noop {} {\bibfield  {journal} {\bibinfo
  {journal} {Neural Networks}\ }\textbf {\bibinfo {volume} {15}},\ \bibinfo
  {pages} {1041--1058} (\bibinfo {year} {2002})}\BibitemShut {NoStop}%
\bibitem [{\citenamefont {Raissi}, \citenamefont {Perdikaris},\ and\
  \citenamefont {Karniadakis}(2019)}]{raissi2019physics}%
  \BibitemOpen
  \bibfield  {author} {\bibinfo {author} {\bibfnamefont {M.}~\bibnamefont
  {Raissi}}, \bibinfo {author} {\bibfnamefont {P.}~\bibnamefont {Perdikaris}},
  \ and\ \bibinfo {author} {\bibfnamefont {G.~E.}\ \bibnamefont
  {Karniadakis}},\ }\bibfield  {title} {\enquote {\bibinfo {title}
  {Physics-informed neural networks: A deep learning framework for solving
  forward and inverse problems involving nonlinear partial differential
  equations},}\ }\href@noop {} {\bibfield  {journal} {\bibinfo  {journal}
  {Journal of Computational Physics}\ }\textbf {\bibinfo {volume} {378}},\
  \bibinfo {pages} {686--707} (\bibinfo {year} {2019})}\BibitemShut {NoStop}%
\bibitem [{\citenamefont {Zhu}\ \emph {et~al.}(2019)\citenamefont {Zhu},
  \citenamefont {Zabaras}, \citenamefont {Koutsourelakis},\ and\ \citenamefont
  {Perdikaris}}]{zhu2019physics}%
  \BibitemOpen
  \bibfield  {author} {\bibinfo {author} {\bibfnamefont {Y.}~\bibnamefont
  {Zhu}}, \bibinfo {author} {\bibfnamefont {N.}~\bibnamefont {Zabaras}},
  \bibinfo {author} {\bibfnamefont {P.-S.}\ \bibnamefont {Koutsourelakis}}, \
  and\ \bibinfo {author} {\bibfnamefont {P.}~\bibnamefont {Perdikaris}},\
  }\bibfield  {title} {\enquote {\bibinfo {title} {Physics-constrained deep
  learning for high-dimensional surrogate modeling and uncertainty
  quantification without labeled data},}\ }\href@noop {} {\bibfield  {journal}
  {\bibinfo  {journal} {Journal of Computational Physics}\ }\textbf {\bibinfo
  {volume} {394}},\ \bibinfo {pages} {56--81} (\bibinfo {year}
  {2019})}\BibitemShut {NoStop}%
\bibitem [{\citenamefont {Swischuk}\ \emph {et~al.}(2019)\citenamefont
  {Swischuk}, \citenamefont {Mainini}, \citenamefont {Peherstorfer},\ and\
  \citenamefont {Willcox}}]{swischuk2019projection}%
  \BibitemOpen
  \bibfield  {author} {\bibinfo {author} {\bibfnamefont {R.}~\bibnamefont
  {Swischuk}}, \bibinfo {author} {\bibfnamefont {L.}~\bibnamefont {Mainini}},
  \bibinfo {author} {\bibfnamefont {B.}~\bibnamefont {Peherstorfer}}, \ and\
  \bibinfo {author} {\bibfnamefont {K.}~\bibnamefont {Willcox}},\ }\bibfield
  {title} {\enquote {\bibinfo {title} {Projection-based model reduction:
  Formulations for physics-based machine learning},}\ }\href@noop {} {\bibfield
   {journal} {\bibinfo  {journal} {Computers \& Fluids}\ }\textbf {\bibinfo
  {volume} {179}},\ \bibinfo {pages} {704--717} (\bibinfo {year}
  {2019})}\BibitemShut {NoStop}%
\bibitem [{\citenamefont {Sekar}\ \emph {et~al.}(2019)\citenamefont {Sekar},
  \citenamefont {Jiang}, \citenamefont {Shu},\ and\ \citenamefont
  {Khoo}}]{sekar2019fast}%
  \BibitemOpen
  \bibfield  {author} {\bibinfo {author} {\bibfnamefont {V.}~\bibnamefont
  {Sekar}}, \bibinfo {author} {\bibfnamefont {Q.}~\bibnamefont {Jiang}},
  \bibinfo {author} {\bibfnamefont {C.}~\bibnamefont {Shu}}, \ and\ \bibinfo
  {author} {\bibfnamefont {B.~C.}\ \bibnamefont {Khoo}},\ }\bibfield  {title}
  {\enquote {\bibinfo {title} {Fast flow field prediction over airfoils using
  deep learning approach},}\ }\href@noop {} {\bibfield  {journal} {\bibinfo
  {journal} {Physics of Fluids}\ }\textbf {\bibinfo {volume} {31}},\ \bibinfo
  {pages} {057103} (\bibinfo {year} {2019})}\BibitemShut {NoStop}%
\bibitem [{\citenamefont {Beck}, \citenamefont {Flad},\ and\ \citenamefont
  {Munz}(2019)}]{beck2019deep}%
  \BibitemOpen
  \bibfield  {author} {\bibinfo {author} {\bibfnamefont {A.}~\bibnamefont
  {Beck}}, \bibinfo {author} {\bibfnamefont {D.}~\bibnamefont {Flad}}, \ and\
  \bibinfo {author} {\bibfnamefont {C.-D.}\ \bibnamefont {Munz}},\ }\bibfield
  {title} {\enquote {\bibinfo {title} {{Deep neural networks for data-driven
  LES closure models}},}\ }\href@noop {} {\bibfield  {journal} {\bibinfo
  {journal} {Journal of Computational Physics}\ }\textbf {\bibinfo {volume}
  {398}},\ \bibinfo {pages} {108910} (\bibinfo {year} {2019})}\BibitemShut
  {NoStop}%
\bibitem [{\citenamefont {Rahman}, \citenamefont {Rasheed},\ and\ \citenamefont
  {San}(2018)}]{rahman2018hybrid}%
  \BibitemOpen
  \bibfield  {author} {\bibinfo {author} {\bibfnamefont {S.}~\bibnamefont
  {Rahman}}, \bibinfo {author} {\bibfnamefont {A.}~\bibnamefont {Rasheed}}, \
  and\ \bibinfo {author} {\bibfnamefont {O.}~\bibnamefont {San}},\ }\bibfield
  {title} {\enquote {\bibinfo {title} {A hybrid analytics paradigm combining
  physics-based modeling and data-driven modeling to accelerate incompressible
  flow solvers},}\ }\href@noop {} {\bibfield  {journal} {\bibinfo  {journal}
  {Fluids}\ }\textbf {\bibinfo {volume} {3}},\ \bibinfo {pages} {50} (\bibinfo
  {year} {2018})}\BibitemShut {NoStop}%
\bibitem [{\citenamefont {Maulik}\ \emph
  {et~al.}(2019{\natexlab{a}})\citenamefont {Maulik}, \citenamefont {San},
  \citenamefont {Jacob},\ and\ \citenamefont {Crick}}]{maulik2019sub}%
  \BibitemOpen
  \bibfield  {author} {\bibinfo {author} {\bibfnamefont {R.}~\bibnamefont
  {Maulik}}, \bibinfo {author} {\bibfnamefont {O.}~\bibnamefont {San}},
  \bibinfo {author} {\bibfnamefont {J.~D.}\ \bibnamefont {Jacob}}, \ and\
  \bibinfo {author} {\bibfnamefont {C.}~\bibnamefont {Crick}},\ }\bibfield
  {title} {\enquote {\bibinfo {title} {Sub-grid scale model classification and
  blending through deep learning},}\ }\href@noop {} {\bibfield  {journal}
  {\bibinfo  {journal} {Journal of Fluid Mechanics}\ }\textbf {\bibinfo
  {volume} {870}},\ \bibinfo {pages} {784--812} (\bibinfo {year}
  {2019}{\natexlab{a}})}\BibitemShut {NoStop}%
\bibitem [{\citenamefont {Maulik}\ \emph
  {et~al.}(2019{\natexlab{b}})\citenamefont {Maulik}, \citenamefont {San},
  \citenamefont {Rasheed},\ and\ \citenamefont {Vedula}}]{maulik2019subgrid}%
  \BibitemOpen
  \bibfield  {author} {\bibinfo {author} {\bibfnamefont {R.}~\bibnamefont
  {Maulik}}, \bibinfo {author} {\bibfnamefont {O.}~\bibnamefont {San}},
  \bibinfo {author} {\bibfnamefont {A.}~\bibnamefont {Rasheed}}, \ and\
  \bibinfo {author} {\bibfnamefont {P.}~\bibnamefont {Vedula}},\ }\bibfield
  {title} {\enquote {\bibinfo {title} {Subgrid modelling for two-dimensional
  turbulence using neural networks},}\ }\href@noop {} {\bibfield  {journal}
  {\bibinfo  {journal} {Journal of Fluid Mechanics}\ }\textbf {\bibinfo
  {volume} {858}},\ \bibinfo {pages} {122--144} (\bibinfo {year}
  {2019}{\natexlab{b}})}\BibitemShut {NoStop}%
\bibitem [{\citenamefont {Pan}\ and\ \citenamefont
  {Duraisamy}(2020)}]{pan2020physics}%
  \BibitemOpen
  \bibfield  {author} {\bibinfo {author} {\bibfnamefont {S.}~\bibnamefont
  {Pan}}\ and\ \bibinfo {author} {\bibfnamefont {K.}~\bibnamefont
  {Duraisamy}},\ }\bibfield  {title} {\enquote {\bibinfo {title}
  {Physics-informed probabilistic learning of linear embeddings of nonlinear
  dynamics with guaranteed stability},}\ }\href@noop {} {\bibfield  {journal}
  {\bibinfo  {journal} {SIAM Journal on Applied Dynamical Systems}\ }\textbf
  {\bibinfo {volume} {19}},\ \bibinfo {pages} {480--509} (\bibinfo {year}
  {2020})}\BibitemShut {NoStop}%
\bibitem [{\citenamefont {Mohan}\ \emph {et~al.}(2020)\citenamefont {Mohan},
  \citenamefont {Lubbers}, \citenamefont {Livescu},\ and\ \citenamefont
  {Chertkov}}]{mohan2020embedding}%
  \BibitemOpen
  \bibfield  {author} {\bibinfo {author} {\bibfnamefont {A.~T.}\ \bibnamefont
  {Mohan}}, \bibinfo {author} {\bibfnamefont {N.}~\bibnamefont {Lubbers}},
  \bibinfo {author} {\bibfnamefont {D.}~\bibnamefont {Livescu}}, \ and\
  \bibinfo {author} {\bibfnamefont {M.}~\bibnamefont {Chertkov}},\ }\bibfield
  {title} {\enquote {\bibinfo {title} {Embedding hard physical constraints in
  neural network coarse-graining of {3D} turbulence},}\ }\href@noop {}
  {\bibfield  {journal} {\bibinfo  {journal} {arXiv preprint arXiv:2002.00021}\
  } (\bibinfo {year} {2020})}\BibitemShut {NoStop}%
\bibitem [{\citenamefont {Muralidhar}\ \emph {et~al.}(2020)\citenamefont
  {Muralidhar}, \citenamefont {Bu}, \citenamefont {Cao}, \citenamefont {He},
  \citenamefont {Ramakrishnan}, \citenamefont {Tafti},\ and\ \citenamefont
  {Karpatne}}]{muralidhar2020physics}%
  \BibitemOpen
  \bibfield  {author} {\bibinfo {author} {\bibfnamefont {N.}~\bibnamefont
  {Muralidhar}}, \bibinfo {author} {\bibfnamefont {J.}~\bibnamefont {Bu}},
  \bibinfo {author} {\bibfnamefont {Z.}~\bibnamefont {Cao}}, \bibinfo {author}
  {\bibfnamefont {L.}~\bibnamefont {He}}, \bibinfo {author} {\bibfnamefont
  {N.}~\bibnamefont {Ramakrishnan}}, \bibinfo {author} {\bibfnamefont
  {D.}~\bibnamefont {Tafti}}, \ and\ \bibinfo {author} {\bibfnamefont
  {A.}~\bibnamefont {Karpatne}},\ }\bibfield  {title} {\enquote {\bibinfo
  {title} {Physics-guided deep learning for drag force prediction in dense
  fluid-particulate systems},}\ }\href@noop {} {\bibfield  {journal} {\bibinfo
  {journal} {Big Data}\ }\textbf {\bibinfo {volume} {8}},\ \bibinfo {pages}
  {431--449} (\bibinfo {year} {2020})}\BibitemShut {NoStop}%
\bibitem [{\citenamefont {Qian}\ \emph {et~al.}(2020)\citenamefont {Qian},
  \citenamefont {Kramer}, \citenamefont {Peherstorfer},\ and\ \citenamefont
  {Willcox}}]{qian2020lift}%
  \BibitemOpen
  \bibfield  {author} {\bibinfo {author} {\bibfnamefont {E.}~\bibnamefont
  {Qian}}, \bibinfo {author} {\bibfnamefont {B.}~\bibnamefont {Kramer}},
  \bibinfo {author} {\bibfnamefont {B.}~\bibnamefont {Peherstorfer}}, \ and\
  \bibinfo {author} {\bibfnamefont {K.}~\bibnamefont {Willcox}},\ }\bibfield
  {title} {\enquote {\bibinfo {title} {{Lift \& Learn: Physics-informed machine
  learning for large-scale nonlinear dynamical systems}},}\ }\href@noop {}
  {\bibfield  {journal} {\bibinfo  {journal} {Physica D: Nonlinear Phenomena}\
  }\textbf {\bibinfo {volume} {406}},\ \bibinfo {pages} {132401} (\bibinfo
  {year} {2020})}\BibitemShut {NoStop}%
\bibitem [{\citenamefont {Pawar}\ \emph
  {et~al.}(2020{\natexlab{a}})\citenamefont {Pawar}, \citenamefont {Ahmed},
  \citenamefont {San},\ and\ \citenamefont {Rasheed}}]{pawar2020data}%
  \BibitemOpen
  \bibfield  {author} {\bibinfo {author} {\bibfnamefont {S.}~\bibnamefont
  {Pawar}}, \bibinfo {author} {\bibfnamefont {S.~E.}\ \bibnamefont {Ahmed}},
  \bibinfo {author} {\bibfnamefont {O.}~\bibnamefont {San}}, \ and\ \bibinfo
  {author} {\bibfnamefont {A.}~\bibnamefont {Rasheed}},\ }\bibfield  {title}
  {\enquote {\bibinfo {title} {Data-driven recovery of hidden physics in
  reduced order modeling of fluid flows},}\ }\href@noop {} {\bibfield
  {journal} {\bibinfo  {journal} {Physics of Fluids}\ }\textbf {\bibinfo
  {volume} {32}},\ \bibinfo {pages} {036602} (\bibinfo {year}
  {2020}{\natexlab{a}})}\BibitemShut {NoStop}%
\bibitem [{\citenamefont {Ahmed}\ \emph
  {et~al.}(2020{\natexlab{a}})\citenamefont {Ahmed}, \citenamefont {San},
  \citenamefont {Rasheed},\ and\ \citenamefont {Iliescu}}]{ahmed2020long}%
  \BibitemOpen
  \bibfield  {author} {\bibinfo {author} {\bibfnamefont {S.~E.}\ \bibnamefont
  {Ahmed}}, \bibinfo {author} {\bibfnamefont {O.}~\bibnamefont {San}}, \bibinfo
  {author} {\bibfnamefont {A.}~\bibnamefont {Rasheed}}, \ and\ \bibinfo
  {author} {\bibfnamefont {T.}~\bibnamefont {Iliescu}},\ }\bibfield  {title}
  {\enquote {\bibinfo {title} {A long short-term memory embedding for hybrid
  uplifted reduced order models},}\ }\href@noop {} {\bibfield  {journal}
  {\bibinfo  {journal} {Physica D: Nonlinear Phenomena}\ ,\ \bibinfo {pages}
  {132471}} (\bibinfo {year} {2020}{\natexlab{a}})}\BibitemShut {NoStop}%
\bibitem [{\citenamefont {Ahmed}\ \emph
  {et~al.}(2020{\natexlab{b}})\citenamefont {Ahmed}, \citenamefont {Pawar},
  \citenamefont {San}, \citenamefont {Rasheed},\ and\ \citenamefont
  {Tabib}}]{ahmed2020nudged}%
  \BibitemOpen
  \bibfield  {author} {\bibinfo {author} {\bibfnamefont {S.}~\bibnamefont
  {Ahmed}}, \bibinfo {author} {\bibfnamefont {S.}~\bibnamefont {Pawar}},
  \bibinfo {author} {\bibfnamefont {O.}~\bibnamefont {San}}, \bibinfo {author}
  {\bibfnamefont {A.}~\bibnamefont {Rasheed}}, \ and\ \bibinfo {author}
  {\bibfnamefont {M.}~\bibnamefont {Tabib}},\ }\bibfield  {title} {\enquote
  {\bibinfo {title} {A nudged hybrid analysis and modeling approach for
  realtime wake-vortex transport and decay prediction},}\ }\href@noop {}
  {\bibfield  {journal} {\bibinfo  {journal} {arXiv preprint arXiv:2008.03157}\
  } (\bibinfo {year} {2020}{\natexlab{b}})}\BibitemShut {NoStop}%
\bibitem [{\citenamefont {Pawar}\ \emph
  {et~al.}(2020{\natexlab{b}})\citenamefont {Pawar}, \citenamefont {Ahmed},
  \citenamefont {San}, \citenamefont {Rasheed},\ and\ \citenamefont
  {Navon}}]{pawar2020long}%
  \BibitemOpen
  \bibfield  {author} {\bibinfo {author} {\bibfnamefont {S.}~\bibnamefont
  {Pawar}}, \bibinfo {author} {\bibfnamefont {S.~E.}\ \bibnamefont {Ahmed}},
  \bibinfo {author} {\bibfnamefont {O.}~\bibnamefont {San}}, \bibinfo {author}
  {\bibfnamefont {A.}~\bibnamefont {Rasheed}}, \ and\ \bibinfo {author}
  {\bibfnamefont {I.~M.}\ \bibnamefont {Navon}},\ }\bibfield  {title} {\enquote
  {\bibinfo {title} {Long short-term memory embedded nudging schemes for
  nonlinear data assimilation of geophysical flows},}\ }\href@noop {}
  {\bibfield  {journal} {\bibinfo  {journal} {Physics of Fluids}\ }\textbf
  {\bibinfo {volume} {32}},\ \bibinfo {pages} {076606} (\bibinfo {year}
  {2020}{\natexlab{b}})}\BibitemShut {NoStop}%
\bibitem [{\citenamefont {Pawar}, \citenamefont {Ahmed},\ and\ \citenamefont
  {San}(2020)}]{pawar2020interface}%
  \BibitemOpen
  \bibfield  {author} {\bibinfo {author} {\bibfnamefont {S.}~\bibnamefont
  {Pawar}}, \bibinfo {author} {\bibfnamefont {S.~E.}\ \bibnamefont {Ahmed}}, \
  and\ \bibinfo {author} {\bibfnamefont {O.}~\bibnamefont {San}},\ }\bibfield
  {title} {\enquote {\bibinfo {title} {Interface learning in fluid dynamics:
  Statistical inference of closures within micro--macro-coupling models},}\
  }\href@noop {} {\bibfield  {journal} {\bibinfo  {journal} {Physics of
  Fluids}\ }\textbf {\bibinfo {volume} {32}},\ \bibinfo {pages} {091704}
  (\bibinfo {year} {2020})}\BibitemShut {NoStop}%
\bibitem [{\citenamefont {Pawar}\ and\ \citenamefont
  {San}(2020)}]{pawar2020prf}%
  \BibitemOpen
  \bibfield  {author} {\bibinfo {author} {\bibfnamefont {S.}~\bibnamefont
  {Pawar}}\ and\ \bibinfo {author} {\bibfnamefont {O.}~\bibnamefont {San}},\
  }\bibfield  {title} {\enquote {\bibinfo {title} {Data assimilation empowered
  neural network parameterizations for subgrid processes in geophysical
  flows},}\ }\href@noop {} {\bibfield  {journal} {\bibinfo  {journal} {arXiv
  preprint arXiv:2006.08901}\ } (\bibinfo {year} {2020})}\BibitemShut {NoStop}%
\bibitem [{\citenamefont {Ahmed}\ \emph
  {et~al.}(2020{\natexlab{c}})\citenamefont {Ahmed}, \citenamefont {San},
  \citenamefont {Kara}, \citenamefont {Younis},\ and\ \citenamefont
  {Rasheed}}]{ahmed2020interface}%
  \BibitemOpen
  \bibfield  {author} {\bibinfo {author} {\bibfnamefont {S.~E.}\ \bibnamefont
  {Ahmed}}, \bibinfo {author} {\bibfnamefont {O.}~\bibnamefont {San}}, \bibinfo
  {author} {\bibfnamefont {K.}~\bibnamefont {Kara}}, \bibinfo {author}
  {\bibfnamefont {R.}~\bibnamefont {Younis}}, \ and\ \bibinfo {author}
  {\bibfnamefont {A.}~\bibnamefont {Rasheed}},\ }\bibfield  {title} {\enquote
  {\bibinfo {title} {Interface learning of multiphysics and multiscale
  systems},}\ }\href@noop {} {\bibfield  {journal} {\bibinfo  {journal}
  {Physical Review E}\ }\textbf {\bibinfo {volume} {102}},\ \bibinfo {pages}
  {053304} (\bibinfo {year} {2020}{\natexlab{c}})}\BibitemShut {NoStop}%
\bibitem [{\citenamefont {Golub}\ \emph {et~al.}(2006)\citenamefont {Golub},
  \citenamefont {Mahoney}, \citenamefont {Drineas},\ and\ \citenamefont
  {Lim}}]{golub2006bridging}%
  \BibitemOpen
  \bibfield  {author} {\bibinfo {author} {\bibfnamefont {G.~H.}\ \bibnamefont
  {Golub}}, \bibinfo {author} {\bibfnamefont {M.~W.}\ \bibnamefont {Mahoney}},
  \bibinfo {author} {\bibfnamefont {P.}~\bibnamefont {Drineas}}, \ and\
  \bibinfo {author} {\bibfnamefont {L.-H.}\ \bibnamefont {Lim}},\ }\bibfield
  {title} {\enquote {\bibinfo {title} {Bridging the gap between numerical
  linear algebra, theoretical computer science, and data applications},}\
  }\href@noop {} {\bibfield  {journal} {\bibinfo  {journal} {SIAM News}\
  }\textbf {\bibinfo {volume} {39}},\ \bibinfo {pages} {1--3} (\bibinfo {year}
  {2006})}\BibitemShut {NoStop}%
\bibitem [{\citenamefont {Lucia}, \citenamefont {Beran},\ and\ \citenamefont
  {Silva}(2004)}]{lucia2004reduced}%
  \BibitemOpen
  \bibfield  {author} {\bibinfo {author} {\bibfnamefont {D.~J.}\ \bibnamefont
  {Lucia}}, \bibinfo {author} {\bibfnamefont {P.~S.}\ \bibnamefont {Beran}}, \
  and\ \bibinfo {author} {\bibfnamefont {W.~A.}\ \bibnamefont {Silva}},\
  }\bibfield  {title} {\enquote {\bibinfo {title} {Reduced-order modeling: new
  approaches for computational physics},}\ }\href@noop {} {\bibfield  {journal}
  {\bibinfo  {journal} {{Progress in Aerospace Sciences}}\ }\textbf {\bibinfo
  {volume} {40}},\ \bibinfo {pages} {51--117} (\bibinfo {year}
  {2004})}\BibitemShut {NoStop}%
\bibitem [{\citenamefont {Taira}\ \emph {et~al.}(2020)\citenamefont {Taira},
  \citenamefont {Hemati}, \citenamefont {Brunton}, \citenamefont {Sun},
  \citenamefont {Duraisamy}, \citenamefont {Bagheri}, \citenamefont {Dawson},\
  and\ \citenamefont {Yeh}}]{taira2019modal}%
  \BibitemOpen
  \bibfield  {author} {\bibinfo {author} {\bibfnamefont {K.}~\bibnamefont
  {Taira}}, \bibinfo {author} {\bibfnamefont {M.~S.}\ \bibnamefont {Hemati}},
  \bibinfo {author} {\bibfnamefont {S.~L.}\ \bibnamefont {Brunton}}, \bibinfo
  {author} {\bibfnamefont {Y.}~\bibnamefont {Sun}}, \bibinfo {author}
  {\bibfnamefont {K.}~\bibnamefont {Duraisamy}}, \bibinfo {author}
  {\bibfnamefont {S.}~\bibnamefont {Bagheri}}, \bibinfo {author} {\bibfnamefont
  {S.~T.}\ \bibnamefont {Dawson}}, \ and\ \bibinfo {author} {\bibfnamefont
  {C.-A.}\ \bibnamefont {Yeh}},\ }\bibfield  {title} {\enquote {\bibinfo
  {title} {Modal analysis of fluid flows: Applications and outlook},}\
  }\href@noop {} {\bibfield  {journal} {\bibinfo  {journal} {AIAA Journal}\
  }\textbf {\bibinfo {volume} {58}},\ \bibinfo {pages} {998--1022} (\bibinfo
  {year} {2020})}\BibitemShut {NoStop}%
\bibitem [{\citenamefont {Fonn}\ \emph {et~al.}(2019)\citenamefont {Fonn},
  \citenamefont {van Brummelen}, \citenamefont {Kvamsdal},\ and\ \citenamefont
  {Rasheed}}]{fonn2019fast}%
  \BibitemOpen
  \bibfield  {author} {\bibinfo {author} {\bibfnamefont {E.}~\bibnamefont
  {Fonn}}, \bibinfo {author} {\bibfnamefont {H.}~\bibnamefont {van Brummelen}},
  \bibinfo {author} {\bibfnamefont {T.}~\bibnamefont {Kvamsdal}}, \ and\
  \bibinfo {author} {\bibfnamefont {A.}~\bibnamefont {Rasheed}},\ }\bibfield
  {title} {\enquote {\bibinfo {title} {Fast divergence-conforming reduced basis
  methods for steady {Navier--Stokes} flow},}\ }\href@noop {} {\bibfield
  {journal} {\bibinfo  {journal} {Computer Methods in Applied Mechanics and
  Engineering}\ }\textbf {\bibinfo {volume} {346}},\ \bibinfo {pages}
  {486--512} (\bibinfo {year} {2019})}\BibitemShut {NoStop}%
\bibitem [{\citenamefont {Pathak}\ \emph {et~al.}(2018)\citenamefont {Pathak},
  \citenamefont {Hunt}, \citenamefont {Girvan}, \citenamefont {Lu},\ and\
  \citenamefont {Ott}}]{pathak2018model}%
  \BibitemOpen
  \bibfield  {author} {\bibinfo {author} {\bibfnamefont {J.}~\bibnamefont
  {Pathak}}, \bibinfo {author} {\bibfnamefont {B.}~\bibnamefont {Hunt}},
  \bibinfo {author} {\bibfnamefont {M.}~\bibnamefont {Girvan}}, \bibinfo
  {author} {\bibfnamefont {Z.}~\bibnamefont {Lu}}, \ and\ \bibinfo {author}
  {\bibfnamefont {E.}~\bibnamefont {Ott}},\ }\bibfield  {title} {\enquote
  {\bibinfo {title} {Model-free prediction of large spatiotemporally chaotic
  systems from data: A reservoir computing approach},}\ }\href@noop {}
  {\bibfield  {journal} {\bibinfo  {journal} {Physical Review Letters}\
  }\textbf {\bibinfo {volume} {120}},\ \bibinfo {pages} {024102} (\bibinfo
  {year} {2018})}\BibitemShut {NoStop}%
\bibitem [{\citenamefont {White}, \citenamefont {Ushizima},\ and\ \citenamefont
  {Farhat}(2019)}]{white2019neural}%
  \BibitemOpen
  \bibfield  {author} {\bibinfo {author} {\bibfnamefont {C.}~\bibnamefont
  {White}}, \bibinfo {author} {\bibfnamefont {D.}~\bibnamefont {Ushizima}}, \
  and\ \bibinfo {author} {\bibfnamefont {C.}~\bibnamefont {Farhat}},\
  }\bibfield  {title} {\enquote {\bibinfo {title} {Neural networks predict
  fluid dynamics solutions from tiny datasets},}\ }\href@noop {} {\bibfield
  {journal} {\bibinfo  {journal} {arXiv preprint arXiv:1902.00091}\ } (\bibinfo
  {year} {2019})}\BibitemShut {NoStop}%
\bibitem [{\citenamefont {Geneva}\ and\ \citenamefont
  {Zabaras}(2020)}]{geneva2020modeling}%
  \BibitemOpen
  \bibfield  {author} {\bibinfo {author} {\bibfnamefont {N.}~\bibnamefont
  {Geneva}}\ and\ \bibinfo {author} {\bibfnamefont {N.}~\bibnamefont
  {Zabaras}},\ }\bibfield  {title} {\enquote {\bibinfo {title} {Modeling the
  dynamics of {PDE} systems with physics-constrained deep auto-regressive
  networks},}\ }\href@noop {} {\bibfield  {journal} {\bibinfo  {journal}
  {Journal of Computational Physics}\ }\textbf {\bibinfo {volume} {403}},\
  \bibinfo {pages} {109056} (\bibinfo {year} {2020})}\BibitemShut {NoStop}%
\bibitem [{\citenamefont {Maulik}\ \emph {et~al.}(2020)\citenamefont {Maulik},
  \citenamefont {Fukami}, \citenamefont {Ramachandra}, \citenamefont
  {Fukagata},\ and\ \citenamefont {Taira}}]{maulik2020probabilistic}%
  \BibitemOpen
  \bibfield  {author} {\bibinfo {author} {\bibfnamefont {R.}~\bibnamefont
  {Maulik}}, \bibinfo {author} {\bibfnamefont {K.}~\bibnamefont {Fukami}},
  \bibinfo {author} {\bibfnamefont {N.}~\bibnamefont {Ramachandra}}, \bibinfo
  {author} {\bibfnamefont {K.}~\bibnamefont {Fukagata}}, \ and\ \bibinfo
  {author} {\bibfnamefont {K.}~\bibnamefont {Taira}},\ }\bibfield  {title}
  {\enquote {\bibinfo {title} {Probabilistic neural networks for fluid flow
  surrogate modeling and data recovery},}\ }\href@noop {} {\bibfield  {journal}
  {\bibinfo  {journal} {Physical Review Fluids}\ }\textbf {\bibinfo {volume}
  {5}},\ \bibinfo {pages} {104401} (\bibinfo {year} {2020})}\BibitemShut
  {NoStop}%
\bibitem [{\citenamefont {Zha}\ \emph {et~al.}(2007)\citenamefont {Zha},
  \citenamefont {Carroll}, \citenamefont {Paxton}, \citenamefont {Conley},\
  and\ \citenamefont {Wells}}]{zha2007high}%
  \BibitemOpen
  \bibfield  {author} {\bibinfo {author} {\bibfnamefont {G.-C.}\ \bibnamefont
  {Zha}}, \bibinfo {author} {\bibfnamefont {B.~F.}\ \bibnamefont {Carroll}},
  \bibinfo {author} {\bibfnamefont {C.~D.}\ \bibnamefont {Paxton}}, \bibinfo
  {author} {\bibfnamefont {C.~A.}\ \bibnamefont {Conley}}, \ and\ \bibinfo
  {author} {\bibfnamefont {A.}~\bibnamefont {Wells}},\ }\bibfield  {title}
  {\enquote {\bibinfo {title} {High-performance airfoil using coflow jet flow
  control},}\ }\href@noop {} {\bibfield  {journal} {\bibinfo  {journal} {AIAA
  Journal}\ }\textbf {\bibinfo {volume} {45}},\ \bibinfo {pages} {2087--2090}
  (\bibinfo {year} {2007})}\BibitemShut {NoStop}%
\bibitem [{\citenamefont {LeGresley}\ and\ \citenamefont
  {Alonso}(2000)}]{legresley2000airfoil}%
  \BibitemOpen
  \bibfield  {author} {\bibinfo {author} {\bibfnamefont {P.}~\bibnamefont
  {LeGresley}}\ and\ \bibinfo {author} {\bibfnamefont {J.}~\bibnamefont
  {Alonso}},\ }\bibfield  {title} {\enquote {\bibinfo {title} {Airfoil design
  optimization using reduced order models based on proper orthogonal
  decomposition},}\ }in\ \href@noop {} {\emph {\bibinfo {booktitle} {Fluids
  2000 conference and exhibit}}}\ (\bibinfo {year} {2000})\ p.\ \bibinfo
  {pages} {2545}\BibitemShut {NoStop}%
\bibitem [{\citenamefont {Zhang}, \citenamefont {Sung},\ and\ \citenamefont
  {Mavris}(2018)}]{zhang2018application}%
  \BibitemOpen
  \bibfield  {author} {\bibinfo {author} {\bibfnamefont {Y.}~\bibnamefont
  {Zhang}}, \bibinfo {author} {\bibfnamefont {W.~J.}\ \bibnamefont {Sung}}, \
  and\ \bibinfo {author} {\bibfnamefont {D.~N.}\ \bibnamefont {Mavris}},\
  }\bibfield  {title} {\enquote {\bibinfo {title} {Application of convolutional
  neural network to predict airfoil lift coefficient},}\ }in\ \href@noop {}
  {\emph {\bibinfo {booktitle} {2018 AIAA/ASCE/AHS/ASC Structures, Structural
  Dynamics, and Materials Conference}}}\ (\bibinfo {year} {2018})\ p.\ \bibinfo
  {pages} {1903}\BibitemShut {NoStop}%
\bibitem [{\citenamefont {Bhatnagar}\ \emph {et~al.}(2019)\citenamefont
  {Bhatnagar}, \citenamefont {Afshar}, \citenamefont {Pan}, \citenamefont
  {Duraisamy},\ and\ \citenamefont {Kaushik}}]{bhatnagar2019prediction}%
  \BibitemOpen
  \bibfield  {author} {\bibinfo {author} {\bibfnamefont {S.}~\bibnamefont
  {Bhatnagar}}, \bibinfo {author} {\bibfnamefont {Y.}~\bibnamefont {Afshar}},
  \bibinfo {author} {\bibfnamefont {S.}~\bibnamefont {Pan}}, \bibinfo {author}
  {\bibfnamefont {K.}~\bibnamefont {Duraisamy}}, \ and\ \bibinfo {author}
  {\bibfnamefont {S.}~\bibnamefont {Kaushik}},\ }\bibfield  {title} {\enquote
  {\bibinfo {title} {Prediction of aerodynamic flow fields using convolutional
  neural networks},}\ }\href@noop {} {\bibfield  {journal} {\bibinfo  {journal}
  {Computational Mechanics}\ }\textbf {\bibinfo {volume} {64}},\ \bibinfo
  {pages} {525--545} (\bibinfo {year} {2019})}\BibitemShut {NoStop}%
\bibitem [{\citenamefont {Rajaram}\ \emph {et~al.}(2020)\citenamefont
  {Rajaram}, \citenamefont {Puranik}, \citenamefont {Renganathan},
  \citenamefont {Sung}, \citenamefont {Pinon-Fischer}, \citenamefont {Mavris},\
  and\ \citenamefont {Ramamurthy}}]{rajaram2020deep}%
  \BibitemOpen
  \bibfield  {author} {\bibinfo {author} {\bibfnamefont {D.}~\bibnamefont
  {Rajaram}}, \bibinfo {author} {\bibfnamefont {T.~G.}\ \bibnamefont
  {Puranik}}, \bibinfo {author} {\bibfnamefont {A.}~\bibnamefont
  {Renganathan}}, \bibinfo {author} {\bibfnamefont {W.~J.}\ \bibnamefont
  {Sung}}, \bibinfo {author} {\bibfnamefont {O.~J.}\ \bibnamefont
  {Pinon-Fischer}}, \bibinfo {author} {\bibfnamefont {D.~N.}\ \bibnamefont
  {Mavris}}, \ and\ \bibinfo {author} {\bibfnamefont {A.}~\bibnamefont
  {Ramamurthy}},\ }\bibfield  {title} {\enquote {\bibinfo {title} {Deep
  {Gaussian} process enabled surrogate models for aerodynamic flows},}\ }in\
  \href@noop {} {\emph {\bibinfo {booktitle} {AIAA Scitech 2020 Forum}}}\
  (\bibinfo {year} {2020})\ p.\ \bibinfo {pages} {1640}\BibitemShut {NoStop}%
\bibitem [{\citenamefont {Drela}(1989)}]{drela1989xfoil}%
  \BibitemOpen
  \bibfield  {author} {\bibinfo {author} {\bibfnamefont {M.}~\bibnamefont
  {Drela}},\ }\bibfield  {title} {\enquote {\bibinfo {title} {{XFOIL: An
  analysis and design system for low Reynolds number airfoils}},}\ }in\
  \href@noop {} {\emph {\bibinfo {booktitle} {Low Reynolds number
  aerodynamics}}}\ (\bibinfo  {publisher} {Springer},\ \bibinfo {year} {1989})\
  pp.\ \bibinfo {pages} {1--12}\BibitemShut {NoStop}%
\bibitem [{\citenamefont {Hess}(1990)}]{hess1990panel}%
  \BibitemOpen
  \bibfield  {author} {\bibinfo {author} {\bibfnamefont {J.~L.}\ \bibnamefont
  {Hess}},\ }\bibfield  {title} {\enquote {\bibinfo {title} {Panel methods in
  computational fluid dynamics},}\ }\href@noop {} {\bibfield  {journal}
  {\bibinfo  {journal} {Annual Review of Fluid Mechanics}\ }\textbf {\bibinfo
  {volume} {22}},\ \bibinfo {pages} {255--274} (\bibinfo {year}
  {1990})}\BibitemShut {NoStop}%
\bibitem [{\citenamefont {Tibshirani}(1996)}]{tibshirani1996comparison}%
  \BibitemOpen
  \bibfield  {author} {\bibinfo {author} {\bibfnamefont {R.}~\bibnamefont
  {Tibshirani}},\ }\bibfield  {title} {\enquote {\bibinfo {title} {A comparison
  of some error estimates for neural network models},}\ }\href@noop {}
  {\bibfield  {journal} {\bibinfo  {journal} {Neural Computation}\ }\textbf
  {\bibinfo {volume} {8}},\ \bibinfo {pages} {152--163} (\bibinfo {year}
  {1996})}\BibitemShut {NoStop}%
\bibitem [{\citenamefont {Heskes}(1997)}]{heskes1997practical}%
  \BibitemOpen
  \bibfield  {author} {\bibinfo {author} {\bibfnamefont {T.}~\bibnamefont
  {Heskes}},\ }\bibfield  {title} {\enquote {\bibinfo {title} {Practical
  confidence and prediction intervals},}\ }in\ \href@noop {} {\emph {\bibinfo
  {booktitle} {Advances in neural information processing systems}}}\ (\bibinfo
  {year} {1997})\ pp.\ \bibinfo {pages} {176--182}\BibitemShut {NoStop}%
\bibitem [{\citenamefont {Lakshminarayanan}, \citenamefont {Pritzel},\ and\
  \citenamefont {Blundell}(2017)}]{lakshminarayanan2017simple}%
  \BibitemOpen
  \bibfield  {author} {\bibinfo {author} {\bibfnamefont {B.}~\bibnamefont
  {Lakshminarayanan}}, \bibinfo {author} {\bibfnamefont {A.}~\bibnamefont
  {Pritzel}}, \ and\ \bibinfo {author} {\bibfnamefont {C.}~\bibnamefont
  {Blundell}},\ }\bibfield  {title} {\enquote {\bibinfo {title} {Simple and
  scalable predictive uncertainty estimation using deep ensembles},}\ }in\
  \href@noop {} {\emph {\bibinfo {booktitle} {Advances in neural information
  processing systems}}}\ (\bibinfo {year} {2017})\ pp.\ \bibinfo {pages}
  {6402--6413}\BibitemShut {NoStop}%
\bibitem [{\citenamefont {Samorani}(2013)}]{samorani2013wind}%
  \BibitemOpen
  \bibfield  {author} {\bibinfo {author} {\bibfnamefont {M.}~\bibnamefont
  {Samorani}},\ }\bibfield  {title} {\enquote {\bibinfo {title} {The wind farm
  layout optimization problem},}\ }in\ \href@noop {} {\emph {\bibinfo
  {booktitle} {Handbook of wind power systems}}}\ (\bibinfo  {publisher}
  {Springer},\ \bibinfo {year} {2013})\ pp.\ \bibinfo {pages}
  {21--38}\BibitemShut {NoStop}%
\bibitem [{\citenamefont {Epstein}\ \emph {et~al.}(2009)\citenamefont
  {Epstein}, \citenamefont {Jameson}, \citenamefont {Peigin}, \citenamefont
  {Roman}, \citenamefont {Harrison},\ and\ \citenamefont
  {Vassberg}}]{epstein2009comparative}%
  \BibitemOpen
  \bibfield  {author} {\bibinfo {author} {\bibfnamefont {B.}~\bibnamefont
  {Epstein}}, \bibinfo {author} {\bibfnamefont {A.}~\bibnamefont {Jameson}},
  \bibinfo {author} {\bibfnamefont {S.}~\bibnamefont {Peigin}}, \bibinfo
  {author} {\bibfnamefont {D.}~\bibnamefont {Roman}}, \bibinfo {author}
  {\bibfnamefont {N.}~\bibnamefont {Harrison}}, \ and\ \bibinfo {author}
  {\bibfnamefont {J.}~\bibnamefont {Vassberg}},\ }\bibfield  {title} {\enquote
  {\bibinfo {title} {Comparative study of three-dimensional wing drag
  minimization by different optimization techniques},}\ }\href@noop {}
  {\bibfield  {journal} {\bibinfo  {journal} {Journal of Aircraft}\ }\textbf
  {\bibinfo {volume} {46}},\ \bibinfo {pages} {526--541} (\bibinfo {year}
  {2009})}\BibitemShut {NoStop}%
\bibitem [{\citenamefont {Rasheed}, \citenamefont {San},\ and\ \citenamefont
  {Kvamsdal}(2020)}]{rasheed2020digital}%
  \BibitemOpen
  \bibfield  {author} {\bibinfo {author} {\bibfnamefont {A.}~\bibnamefont
  {Rasheed}}, \bibinfo {author} {\bibfnamefont {O.}~\bibnamefont {San}}, \ and\
  \bibinfo {author} {\bibfnamefont {T.}~\bibnamefont {Kvamsdal}},\ }\bibfield
  {title} {\enquote {\bibinfo {title} {Digital twin: Values, challenges and
  enablers from a modeling perspective},}\ }\href@noop {} {\bibfield  {journal}
  {\bibinfo  {journal} {IEEE Access}\ }\textbf {\bibinfo {volume} {8}},\
  \bibinfo {pages} {21980--22012} (\bibinfo {year} {2020})}\BibitemShut
  {NoStop}%
\bibitem [{\citenamefont {Pawar}(2020)}]{githubpgml}%
  \BibitemOpen
  \bibfield  {author} {\bibinfo {author} {\bibfnamefont {S.}~\bibnamefont
  {Pawar}},\ }\href@noop {} {\enquote {\bibinfo {title} {{PGML}},}\ }\bibinfo
  {howpublished} {\url{https://github.com/surajp92/PGML}} (\bibinfo {year}
  {2020})\BibitemShut {NoStop}%
\end{thebibliography}%
\end{document}